\documentclass[letterpaper, 10 pt, conference]{ieeeconf}  % Comment this line out if you need a4paper

\IEEEoverridecommandlockouts                              % This command is only needed if 
\let\labelindent\relax
\usepackage{enumitem}
\usepackage{balance}
\usepackage{usebib}
\usepackage{amsmath,amssymb,amsfonts}
\usepackage{algorithmic}
\usepackage{graphicx}
\usepackage[caption=false, font=footnotesize]{subfig}
\usepackage{textcomp}
\usepackage{xcolor}
\usepackage{hyperref}
\hypersetup{
    colorlinks=true,
    linkcolor=blue,
    filecolor=magenta,      
    urlcolor=cyan,
    }

\newcommand{\vx}{{\bf x}}
\newcommand{\vp}{{\bf p}}
\newcommand{\vv}{{\bf v}}
\newcommand{\vu}{{\bf u}}
\newcommand{\vy}{{\bf y}}
\newcommand{\vF}{{\bf F}}

\title{\LARGE \bf
Dynamics Modeling using Visual Terrain Features for High-Speed Autonomous Off-Road Driving
}

\author{Jason Gibson$^{1}$\textsuperscript{\textsection}, Anoushka Alavilli$^{2}$\textsuperscript{\textsection}, Erica Tevere$^{3}$, Evangelos A. Theodorou$^{1}$, and Patrick Spieler$^{3}$% <-this % stops a space
\thanks{\textsuperscript{\textsection} These authors contributed equally.}
\thanks{$^{1}$Jason Gibson and Evangelos Theodorou are with the Autonomous Control and Decision Systems Lab, Georgia Institute of Technology,
        Atlanta, GA 30332, USA
        {\tt\small \{jgibson37, evangelos.theodorou\}@gatech.edu}}%
\thanks{$^{2}$Anoushka Alavilli is with the Robotics Institute, Carnegie Mellon University,
        Pittsburgh, PA 15213, USA
        {\tt\small apalavil@andrew.cmu.edu}}%
\thanks{$^{3}$Patrick Spieler and Erica Tevere are with the NASA Jet Propulsion Laboratory, California Institute of Technology,
        Pasadena, CA 91125, USA
        {\tt\small \{patrick.spieler, erica.l.tevere\}@jpl.nasa.gov}}%
\thanks{The research was carried out at the Jet Propulsion Laboratory, California Institute of Technology, under a contract with the National Aeronautics and Space Administration (80NM0018D0004). This work was  supported by Defense Advanced Research Projects Agency (DARPA). Approved for Public Release,
Distribution Unlimited. \copyright 2024. All rights reserved.} %
}

\begin{document}

\maketitle
\thispagestyle{empty}
\pagestyle{empty}

\begin{abstract}

Rapid autonomous traversal of unstructured terrain is essential for scenarios such as disaster response, search and rescue,  or planetary exploration.
As a vehicle navigates at the limit of its capabilities over extreme terrain, its dynamics can change suddenly and dramatically. For example, high-speed and varying terrain can affect parameters such as traction, tire slip, and rolling resistance. 
To achieve effective planning in such environments, it is crucial to have a dynamics model that can accurately anticipate these conditions. 
In this work, we present a hybrid model that predicts the changing dynamics induced by the terrain as a function of visual inputs. 
We leverage a pre-trained visual foundation model (VFM) DINOv2, which provides rich features that encode fine-grained semantic information.
To use this dynamics model for planning, we propose an end-to-end training architecture for a projection distance independent feature encoder that compresses the information from the VFM, enabling the creation of a lightweight map of the environment at runtime.
We validate our architecture on an extensive dataset (hundreds of kilometers of aggressive off-road driving) collected across multiple locations as part of the DARPA Robotic Autonomy in Complex Environments with Resiliency (RACER) program. \href{https://www.youtube.com/watch?v=dycTXxEosMk}{https://www.youtube.com/watch?v=dycTXxEosMk}

\end{abstract}

\section{Introduction}

In off-road autonomous driving, varying terrain geometries and properties influence the terradynamics experienced by the vehicle. 
Terrain geometries include obstacles, hills, trenches, and slopes, while terrain properties include stiffness, friction, and other properties that vary with the vegetation, gravel, or sand on the vehicle's path. 
When driving without prior knowledge of the environment, real-time information about the terrain can provide insight into the safety and efficiency of traversing various paths. 
For example, Mars Science Laboratory (the Curiosity rover) experienced unanticipated wheel damage, prompting the choice of path plans involving more benign terrain \cite{msl_wheel_damage}. 
Similarly, the vehicle slip experienced by the Mars Exploration Rovers (Spirit and Opportunity) was difficult to predict given the highly varied terrain \cite{mer_wheel_slip}.

\begin{figure}[!t]
\centering
\includegraphics[width=\linewidth]{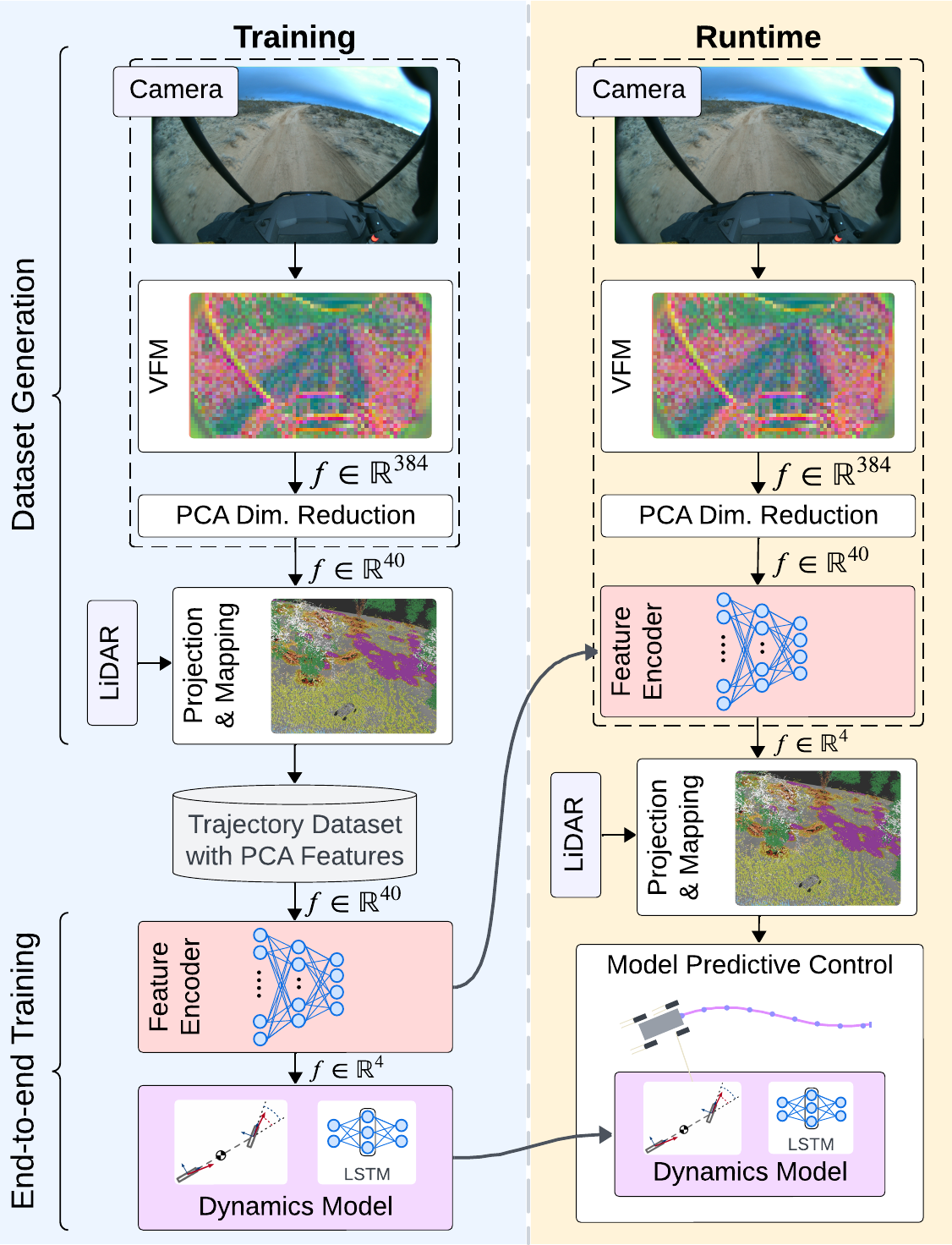}
% \caption{\textcolor{red}{TODO: this is a placeholder--will replace}}
% https://lucid.app/lucidchart/e968623a-4f94-40b7-b6b8-b526f4b5fdcc/edit?invitationId=inv_532d7953-57ab-42bf-a893-49374a6f4c62&page=0_0
\caption{Architecture of dynamics learning with visual features.
A feature encoder is trained end-to-end with the dynamics model on a dataset of high-dimensional visual features from a VFM.
This feature encoder reduces the visual information to a low-dimensional dynamics relevant feature space.
At runtime, it processes features in image space (dashed line) before projection and accumulation in a 3D map. This makes the map aggregation step computationally tractable.
The 3D map is flattened to a top down 2D terrain feature map used by the dynamics model in the MPC planner.}
\label{autonomystack}
\end{figure}

Dynamics modeling for field robots using visual inputs has been an area of research since before visual foundation models (VFMs) were available. For example, in \cite{lagr}, different vehicle slip models were determined for a small number of different terrain conditions (such as grass, gravel, etc.). Though discretizing terrain types could suffice in some environments and when the robot is not operating at its hardware limits, a continuous representation of terrain is more desirable for our purposes, where terrain varies drastically and the driving is highly aggressive. 
%This method suffered from high error rates, and with the recent advances in VFMs, more precise (and continuous) representations of terrain conditions are feasible. 

VFMs are large-scale, pre-trained neural networks designed to handle a wide range of vision-related tasks, such as image classification, segmentation, and feature extraction. VFMs such as DINOv2~\cite{dino} are trained to produce general-purpose features, which could be used for applications such as classification or segmentation. Features can carry semantic meaning; for example, some feature maps may encode depth information and provide linear image segmentation. In this work, we leverage DINOv2 to extract compact and continuous representations of terrain, capturing visual details that can inform the vehicle's dynamics.

\begin{figure}[t]
\centering
    \includegraphics[width=.161\textwidth]{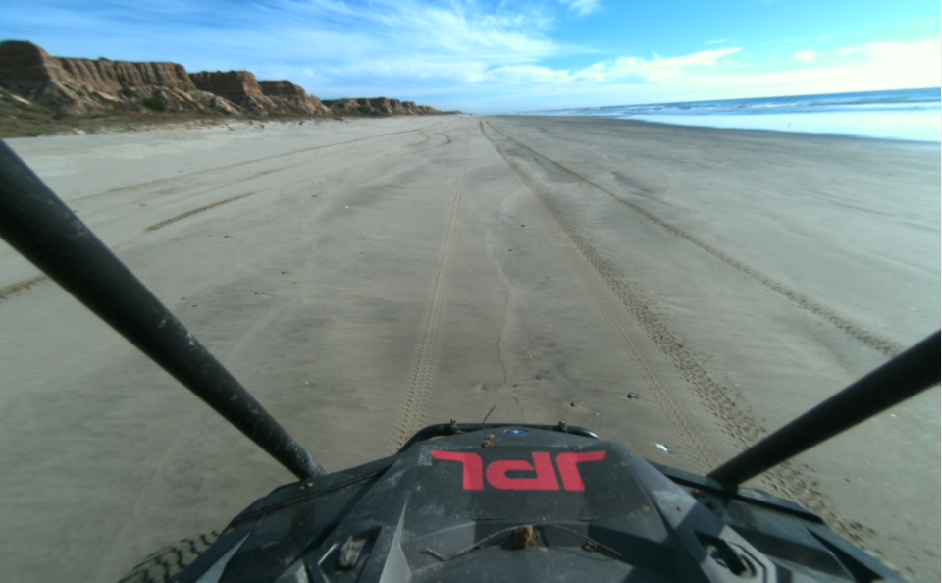}\hfill
    \includegraphics[width=.161\textwidth]{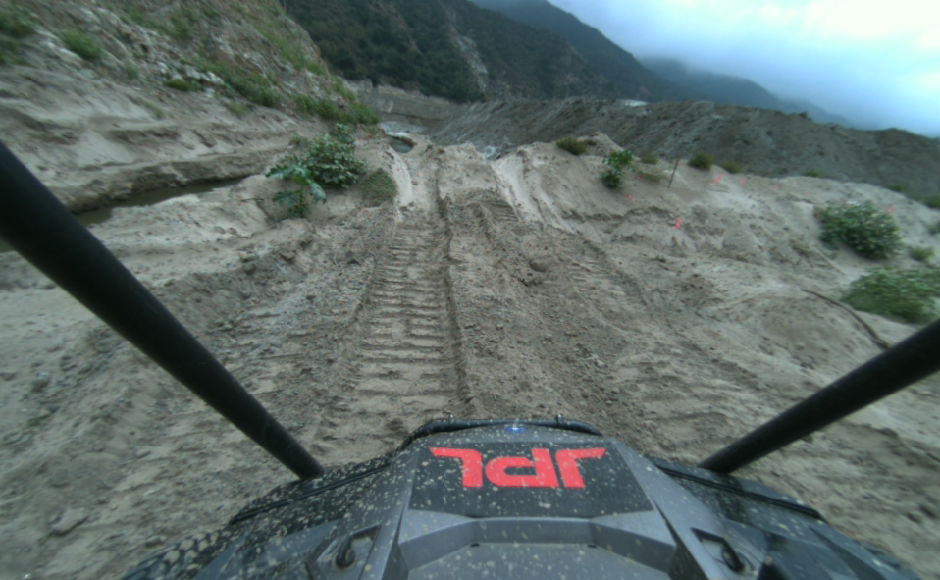}\hfill
    \includegraphics[width=.161\textwidth]{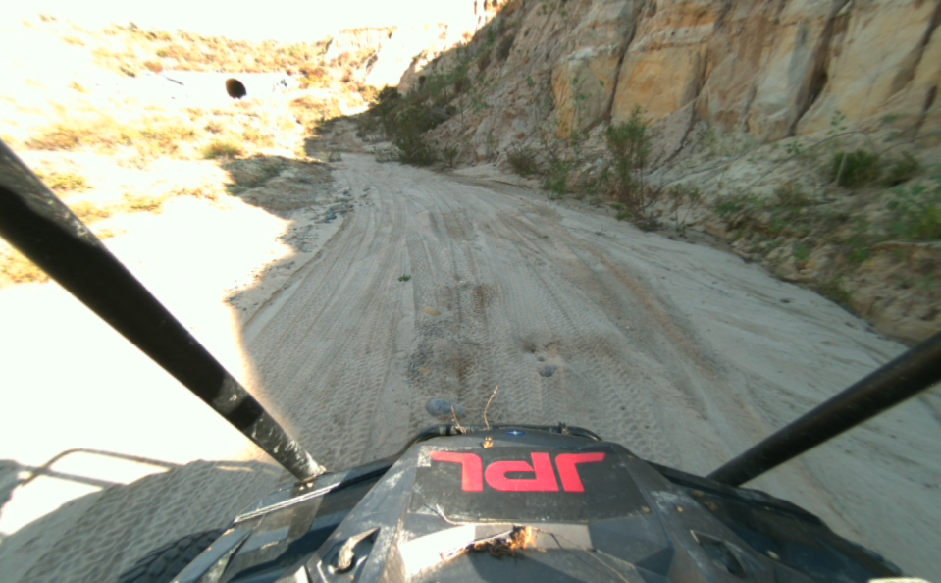}\hfill
    \vspace{0.01cm}
    %\\[\smallskipamount]
    \includegraphics[width=.161\textwidth]{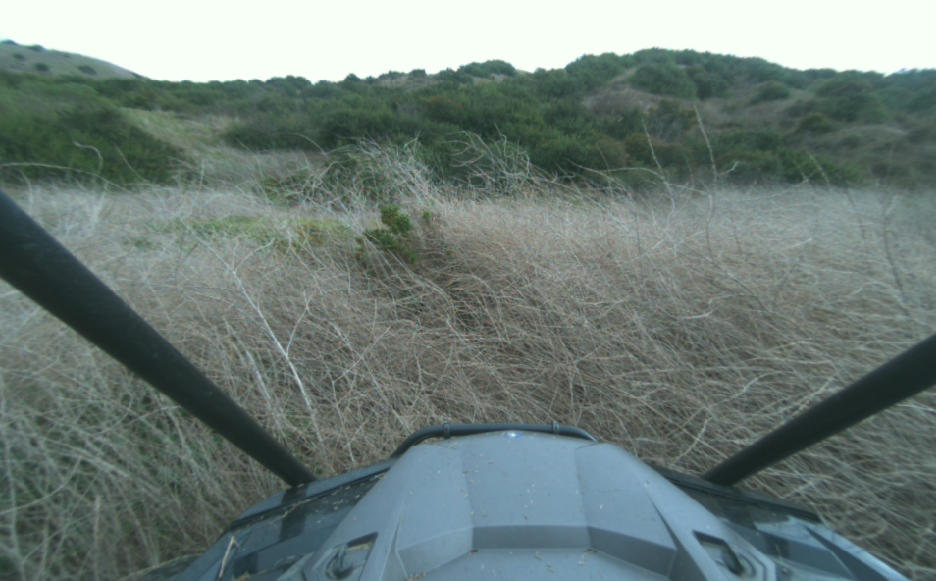}\hfill
    \includegraphics[width=.161\textwidth]{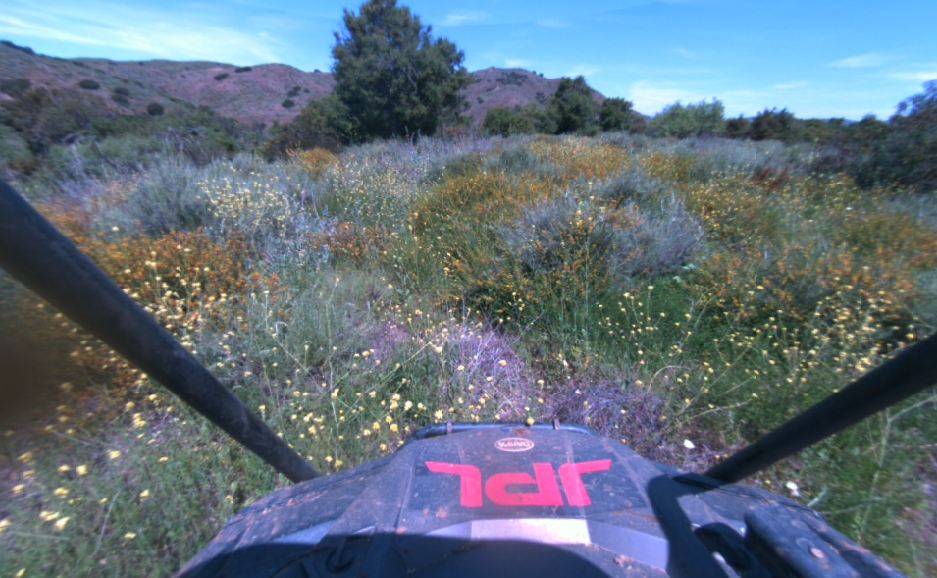}\hfill
    \includegraphics[width=.161\textwidth]{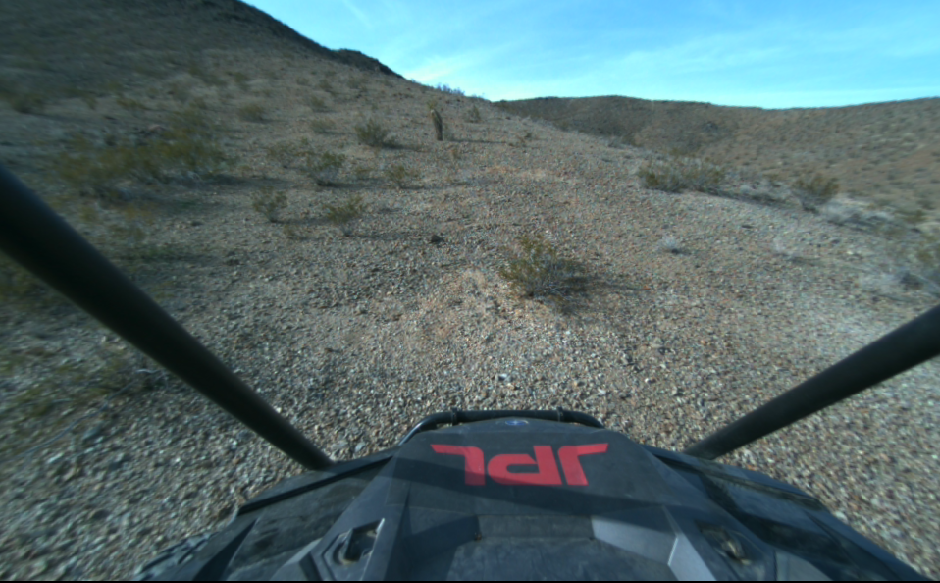}
\caption{Terrain geometries and properties vary significantly across the environments. Images show a selection of diverse terrain (from top left to bottom right: packed sand, muddy ditches and ruts, loose dirt trail, tall grass, dense overgrown vegetation, steep slopes) for which visual inputs of the terrain inform the changing dynamics of the vehicle.}
\label{varied_terrains}
\end{figure}

Understanding the vehicle's surrounding terrain is especially important as the coupling of high speeds and varying ground terrain introduce complex, nonlinear, and time-varying properties in the dynamics of the vehicle as aspects such as the traction, cornering stiffness, and rolling resistance of the vehicle change. For example, the vehicle's dynamics when driving on slippery grass differ from those on dry trails. Examples of terrain where the vehicles dynamics may be affected as a result of the terradynamics of the environment are shown in Fig.~\ref{varied_terrains}.

%\subsection{Contributions}
These challenges highlight the importance of integrating perceptual inputs into our dynamics model. Most importantly, the dependence on visual features will allow for the \textit{anticipation} of changes to vehicle dynamics, ultimately reducing the need for model correction and adaptation as the vehicle drives on varying terrain. To this end, the contributions of this work are as follows:
%\begin{enumerate}[noitemsep,  leftmargin=*]
%\item A hybrid dynamics model (i.e., a model with both physics-based parametric and neural-network components) that predicts changes in terradynamics as a function of terrain using visual features derived from a VFM.
%\item A training architecture that enables end-to-end learning of a feature encoder that can be used to lower the computational burden of tracking high-dimensional features in map space at runtime.
%\item A novel method for compressing a visual feature space that is robust to projection distance and occlusion, enabling generalization to real driving scenarios.
%\item Validation of our method on a large dataset of aggressive off-road driving across a variety of rugged terrains.
%\end{enumerate}
\textbf{1).} A hybrid dynamics model (i.e., a model with both physics-based parametric and neural-network components) that predicts changes in terradynamics as a function of terrain using visual features derived from a VFM.
\textbf{2).} A training architecture that enables end-to-end learning of a feature encoder that can be used to lower the computational burden of tracking high-dimensional features in map space at runtime.
\textbf{3).} A novel method for compressing a visual feature space that is robust to projection distance and occlusion, enabling generalization to real driving scenarios.
\textbf{4).} Validation of our method on a large dataset of aggressive off-road driving across a variety of rugged terrains.

\section{Related Work}
% Structure (each paragraph has: 1)  literature, 2) what we do different 

In off-road autonomy literature, many studies aim to improve the quality of the traversability map by incorporating visual information about the surrounding terrain \cite{bevnet}, \cite{evora}, \cite{roadrunner}, \cite{wayfast}, \cite{wild}, \cite{msr}. For instance, \cite{bevnet} combines semantic labels and geometric hazard identification to determine the costs in the traversability map. \cite{evora} learns a traction model to estimate the slip parameters of the terrain and factors this into the cost of the traversability map. Some works, such as \cite{msr}, classify terrain into a discrete set of options, like smooth and rough regolith, limiting their approach. The main goal of these methods is to avoid unsafe paths. Less explored is the use of visual terrain features to inform vehicle dynamics modeling, which is necessary for accurate control and planning at a higher resolution.

Other works, such as~\cite{liu2022} and~\cite{kabzan}, also propose hybrid models (i.e., a model that has parametric and learned components) for high-speed vehicle dynamics modeling. In both works, the vehicle drives on-road, and therefore neither approach uses visual terrain features to inform the model.

Most relevant to this research are~\cite{magic_vfm} and~\cite{hutter}. Both methods use the DINO or DINOv2 VFM to inform a model about the traversed terrain. 
\cite{hutter} employs the visual features to inform two physical parameters, stiffness and friction, while~\cite{magic_vfm} uses this visual terrain-feature-based model for adaptive control.
Compared to~\cite{hutter}, our method outputs a richer representation of the terrain by training a feature encoder end-to-end with a dynamics model.
In contrast to~\cite{magic_vfm}, which processes the terrain information at each camera frame and uses that instantaneous information for control, our method incorporates a lower-dimensional representation of the DINOv2 features into a 2D map, which is then queried at each of the wheel locations.
This allows our method to accurately capture the spatial distribution of terrain features and terrain transitions, which is particularly relevant for predicting trajectories in a Model Predictive Control (MPC) framework.

Perception challenges such as lighting conditions, distance bias, and occlusions can all contribute to the lack of complete and immutable perception of surrounding terrain. Several works address the issue of consistency of data from the vehicle's surroundings during driving. For instance,~\cite{jung2024vstrong} proposes a self-supervised learning framework that uses both a VFM and human driving trajectories for terrain traversability learning. Similarly, \cite{xue2023contrastive} uses human driving data and weakly-supervised contrastive learning. Labels of the surrounding terrain may change as the vehicle drives, and previously unseen voxels may be filled in as the vehicle passes through occlusions such as bushes. 
Multimodal mapping approaches like \cite{erni2023mem} have used accumulation strategies such as latest information, exponential averaging and Bayesian updates.
Pyramid occupancy networks have also been used to accumulate map predictions across timesteps. For example, \cite{roddick2020predicting} uses Bayesian filtering to fuse voxel map information over a time buffer of driving. To further analyze these perception challenges, we examine the consistency of the DINOv2 features with respect to the distance at which they are collected and the dimensionality reduction we perform.

\section{Mapping of Visual Features for Controls}

To provide context, we briefly introduce the main components of JPL's RACER autonomy stack used in this work.
A mapping module combines image and LiDAR information by projecting the pointclouds onto image data. This module then aggregates the resulting pointclouds, augmented with the image data, in a 3D voxel map.
Depending on the downstream application, the image data can be represented as semantic class probabilities or as a latent feature space.
This 3D voxel map is further processed into a 2D traversability map, which is used by the planning and control modules.
This 2D traversability map contains multiple layers that encode various quantities such as elevation, obstacles, and planning costs. It also contains terrain features for the dynamics model in this work, as detailed in the following sections.

\subsection{Model Predictive Control with Learned Dynamics}

As in our previous work \cite{multistep}, we learn a dynamics model for our planner, Model Predictive Path Integral (MPPI)~\cite{mppi}, to generate optimal trajectories for the vehicle. At a high level, MPPI is a sampling-based planner, and it operates by sampling various trajectories, performing forward rollouts of the dynamics, and optimizing the trajectories based on the cost of these sampled rollouts. 
MPPI is well-suited for our application since it is easily parallelizable on the GPU \cite{mppi-generic} and supports the use of complex, sparse cost functions. %such as our traversability cost map. 
We also incorporate safety constraints in this framework, such as minimizing rollover risk. 
The specific cost function and the MPPI variation we use are described in \cite{colored_mppi}. The planning at this scale is optimized over a 5-second prediction horizon, requiring accurate and computationally efficient dynamics modeling for effective planning.
The terrain elevation and visual terrain features in the 2D traversability map are queried by the dynamics model at the location of each wheel along the trajectory rollout.

\subsection{Dataset Collection Pipeline}

We use the autonomy stack to generate the dynamics dataset with visual features from recorded driving data.
We use a perception pipeline nearly identical to the one used at runtime, feed it recorded sensor data, and record relevant outputs.
The key difference between the runtime and replay configuration is the placement of the feature encoder, as emphasized in Fig.~\ref{autonomystack}.
The feature encoder, described further in Section~\ref{sec:feature-dynamics-model}, compresses visual features into a low-dimensional, dynamics-relevant feature space.
When creating the dataset, we store a high-dimensional feature vector to train the feature encoder jointly with the dynamics model.
At runtime, the feature encoder runs before projection and mapping, reducing the computation and memory burden of mapping visual features to enable real-time execution.

\subsubsection{Visual Features}

As the vehicle drives, images are captured from four RGB cameras facing forward, back, and to both sides. The front and side cameras operate at a rate of 10Hz, while the rear camera captures images at 2Hz, all with a resolution of $960\times594$ pixels.
The camera images are rectified before being processed by the DINOv2 VFM.
We employ the smallest distilled ViT-S/14 network size which uses an embedding dimension of $\mathbb{R}^{384}$.

\begin{figure}[t]
\centering
\includegraphics[width=0.8\linewidth]{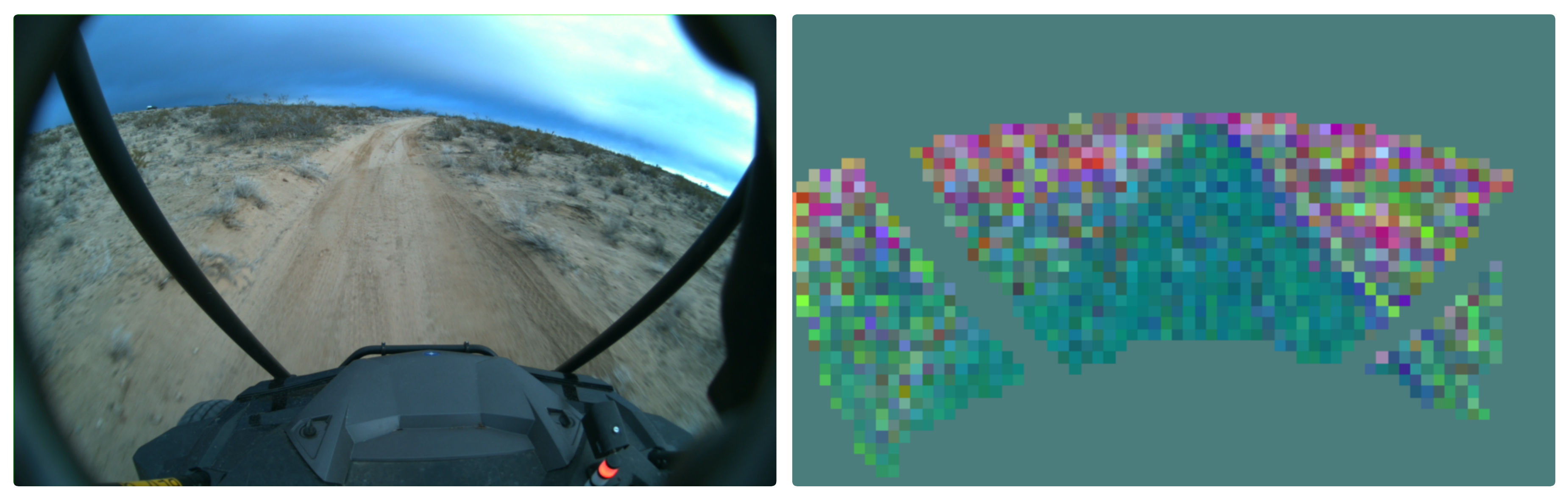}
\caption{Left: A forward-facing image of size $\mathbb{R}^{960\times594\times3}$ (in RGB). Right: VFM output of size $\mathbb{R}^{68\times42\times384}$, where each $14\times14$ pixel patch results in one feature vector of size $\mathbb{R}^{1\times384}$. DINOv2 features from ground regions undergo PCA, and the first three components are visualized in RGB. The result effectively segments on- and off-trail terrain.} %TODO-camera-ready: \textbf{c)} Visualization of the first feature resulting from the learned compression that reduces the features to $\mathbb{R}^{68\times42\times4}$ in our baseline model.}
\label{fwd_facing_dino}
\end{figure}

\subsubsection{Dimensionality Reduction using Principal Component Analysis}

To lower the memory and computational requirements during the dataset generation, we apply Principal Component Analysis (PCA) projection on the output of the VFM (Fig.~\ref{autonomystack}).
To generate the PCA basis, a set of 175 images were manually selected, covering various terrain types and lighting conditions.
These images are run through the VFM to obtain feature images, which are then masked to retain only feature vectors from ground pixels, and a PCA basis is computed from these.
The PCA projection provides a linear map from $\mathbb{R}^{n_{\text{vfm}}} \mapsto \mathbb{R}^{n_{\text{pca}}}$.
The replay can run slower than real time, so it can support a high feature dimension. The size of $n_{\text{pca}}$ is a trade-off between dataset processing speed/memory requirements and model quality.
We choose $n_{\text{pca}} = 40$, which greatly reduces the memory requirements while maintaining a sufficiently rich feature space.
As shown in Section~\ref{sec:results_post_processed} and Fig.~\ref{fig:num_pca_features}, the performance is not highly sensitive to the PCA basis size.
An example image and the corresponding RGB visualization of the first three components of the PCA features are shown in Fig. \ref{fwd_facing_dino}.

\subsubsection{Mapping of Visual Features}\label{3Dto2D}

% - LiDAR projection (leveraging exisitng mapping infrastructure)
% - accumulation strategy closest in voxel space
%     - why this works well for semantics % TODO cite ISRR paper here when on arxiv
% - compression to 2D using lowest
% - infill over xx m 
% - note/repeat that this applies to both dataset generation and runtime, with runtime mapping the output of the feature encoder

The visual features are mapped into a local, robot-centric map, leveraging existing mapping infrastructure.
In the case of offline dataset generation, the features are the output of the PCA compression, whereas at runtime they are the output of the feature encoder, which substantially further compresses the features.
To map these features, LiDAR pointcloud data is projected onto the image plane of the feature image, and the pointcloud values are augmented with the corresponding feature vector.
The pointcloud is then aggregated temporally into a 3D voxel map with 0.2~m resolution.
To fuse multiple measurements, we retain the closest observation, allowing for instant updates and temporal stability when moving away from a location. % TODO cite ISRR paper here when on arxiv
The visual terrain features in the 3D voxel data are compressed to a 2D map output by taking the lowest valid data point within each vertical stack of voxels, under the assumption that the visual feature of the ground, and not anything above, is most relevant for dynamics learning.
To close small gaps of missing data due to LiDAR sparsity or small occlusions, missing data is filled in with data from the nearest neighboring cell in a 0.4~m radius.

\subsubsection{Dataset Extraction}
% - what is in the dataset
%     - features of 4 wheel locations queried into the map
% - hindsight map vs distance based maps
% - mention how hindsight / later maps have rear lidar/camera

To generate a dataset for dynamics learning, the vehicle state and control trajectories are stored together with the visual features and the elevation surface normals under each wheel.
This trajectory is later split up into chunks that form the actual 5~s trajectory prediction dataset.
Visual features may vary with distance, so we store multiple values in the map at different points in time.
In particular, the back camera and LiDAR typically fill in data behind small vegetation.
The maps from which to take features are chosen based on the distance between the query point and the map origin (approximately the latest robot location).
As a result, each trajectory, at every point along it, will have multiple sets of features mapped from various distances, which we use later to train a distance-independent encoder.
In addition, features from a ``hindsight" map, which contains the last valid data (and therefore the least amount of missing data), are stored.

\section{Hybrid Vehicle Dynamics Model}

Our previous work on dynamics modeling, such as \cite{multistep}, has drawn upon hybrid models, which include both physics-based parametric and neural network dynamics components. This approach was chosen primarily for the reliability of the parametric components when the networks are in low data regimes. The vehicle dynamics are divided into four main components: brake, steering, engine, and terradynamics. For the first three modules, we model the delay in actuation or RPM (revolutions per minute) as a state.
In this current work, we model each component with a hybrid model, except for the engine dynamics, for which we predict the RPM of the engine directly from the throttle and speed of the vehicle. A key difference from our previous work is that we compute compensations to predicted forces rather than directly predict the state.
We predict the state vector $\vx \in \mathbb{R}^6$ for the terradynamics model containing the inertial positions $\vp = [p_x, p_y]$, the yaw angle $\phi$, the velocity in the body frame $\vv = [v_x, v_y]$, and the angular rate $r$.
An additional 4 states are predicted, such as pressure for the brake, position and velocity for the steering angle, and a value for the engine RPM, but their models are fixed during training of the terradynamics.
All constants and networks are learned using the Adam optimization algorithm \cite{adam} on training data drawn from a variety of field test sites.

\subsection{Parametric Bicycle Model}

To model the parametric portion of the terradynamics hybrid model, we employ a bicycle dynamics model, largely drawn from~\cite{kabzan} in form. 
We include our own computation of forward force using the predicted engine RPM and commanded throttle.
We modify the Pacejka tire model and yaw rate equation to avoid integrator stability issues.
\begin{align}
    \vF = \begin{bmatrix}
        \left(P\left(x_{rpm}\right) P\left(u_{th}\right) - P\left(x_{br}\right) - \beta\left(v_x\right)\right) \eta_z\\
        \left(D_R\text{sin}\left(C_R\text{tanh}\left(B_R\alpha_R\right)\right)\right) \eta_z\\
        \left(D_F\text{sin}\left(C_F\text{tanh}\left(B_F\alpha_F\right)\right)\right) \eta_z\\
        \left(\frac{v_x}{C_L} \delta\right) C_{r} - C_{r, d} r
    \end{bmatrix},
    \label{eq:force_equations} 
\end{align}
where $\vF = [F_x, F_{yf}, F_{yb}, F_r]$ defines a vector of forces, $x_{rpm}, x_{br}, \delta$ are state variables computed by the delay models (engine, brake, steering), $u_{th}$ is the commanded throttle, $v_x, v_y, r$ are the body velocities and yaw rate respectively, $\eta_z$ is the current normal vector from the elevation map rotated into body frame and averaged over the wheels, and $D, B, C$ etc. are all fit constants.
$P(\cdot)$ is a quadratic polynomial of the input and $\beta(\cdot)$ is a scaled $\tanh$ of the input.
The first equation in~\eqref{eq:force_equations} is the forward force applied to each wheel, the second two are the lateral forces on the tires, and the final equation is an approximation for yaw rate.
The front and rear wheel slip angles, $\alpha_F$ and $\alpha_R$, are shown below:
\begin{align}
    \alpha_R = \text{arctan}\left(\frac{v_y - L_Rr}{\text{max}(C_{max},v_x)}\right),&\label{tire3}\\
    \alpha_F = \text{arctan}\left(\frac{v_y + L_Fr}{\text{max}(C_{max},v_x)}\right) - \delta,& \label{tire4}
\end{align} where $C_{max}, L_R, L_F$ are learned parameters. 
$C_{max}$ controls the stiffness of the slip angle equation, creating a trade-off between the slip angle accuracy and the stiffness.
%The modification was made to account for a negative $v_x$ and to aid in integrator stability when $v_x$ is small.

Finally, the forces are then converted to body frame using geometric transforms $h(\vF, \vx):\mathbb{R}^4\times \mathbb{R}^6 \rightarrow \mathbb{R}^6$ to compute the derivatives of body rates using
\begin{equation}\label{eq:all}
\begin{aligned}
    \dot{v}_x &= \frac{(1 + \cos\delta) F_x - F_{yf} \sin\delta}{m} - C_{x, d} v_x^2  - C_{x, g}\eta_x + v_y r, \\
    \dot{v}_y &= \frac{F_{yb} + \cos\delta F_{yf} + F_x \sin\delta}{m} - C_{y, d} v_y^2  -C_{y, g}\eta_y - v_x r, \\
    \dot{r} &= F_r, \\
    \dot{\vp} &= R(\phi)\vv, \quad
    \dot{\phi} = r,
\end{aligned}
\end{equation}
where $C_{\cdot, d}, C_{\cdot,g}$ are the drag and gravity coefficients, and $m$ is the vehicle mass.
The system in~\eqref{eq:all} is integrated using forward Euler integration with a $\Delta t = 0.02s$.

We predict a compensation of the parametric force $\bf F$ using the neural networks,
\begin{align}
    \dot{\hat{\vx}}_t &= h\!\left(\vF_t + \zeta_\mu\!\left(\hat{\vx}_t, \vu_t, \vy_t, \vF_t\right), \hat{\vx}_t\right),\label{hybridmodelwithoutencoder}
\end{align}
where $\hat{\vx}_t$ is the predicted state at time $t$, $\vu_t, \vy_t$ are the control and map inputs respectively, and $\zeta_\mu$ is an LSTM initialized as in \cite{multistep} over a local horizon of historical values $[t-\tau, t],\tau=0.2$.
We emphasize that the state values $\hat{\vx}_t$ are the predicted ones from delay models and dynamics, while the other inputs are what was actually seen or commanded.
This is the critical component for error correction in the multistep learning approach.

\section{Feature-Based Dynamics Model}
\label{sec:feature-dynamics-model}

Not all DINOv2 features of the surrounding terrain will be relevant to the dynamics of the vehicle. For example, any feature maps that encode lighting conditions or depth information should not be used in dynamics modeling. We incorporate a feature encoder $\zeta_E \in \mathbb{R}^{n_{\text{pca}}} \mapsto \mathbb{R}^{n_{\text{encoder}}}$, therefore, to compress the feature space into a more compact, dynamically-relevant subspace and train the network within the training pipeline to ensure that only the feature information correlated with dynamics is retained.
Each wheel location is compressed individually giving $\mathbb{R}^{4 \cdot n_\text{encoder}}$ features that are used as additional inputs to the neural network.
The feature encoder is a fully-connected neural network that processes the VFM features extracted at each tire location, resulting in a 40-dimensional input layer run $4$ times independently for each wheel. 
It includes hidden layers of size $[64, 32]$ with $\tanh$ activation functions, followed by an output layer of size $n_{\text{encoder}}$, which we vary in \ref{fig:compression_size}.
%We focus on values for $n_\text{encoder}$ that are divisible by 2 or 4 for hardware acceleration in querying them from a CUDA texture for use in MPPI. 
%This also reduces the memory and computation overhead of using a larger feature vectors in the rest of the pipeline.

\subsection{Learning Distance-Independent Compression}
\label{sec:learning_dist_ind_features}

For the terradynamics model, we follow this baseline structure but add an additional network $\zeta_E$ that passes encodings of the DINOv2 terrain features into the input of the LSTM $\zeta_\mu$ as follows:
\begin{align}
    \dot{\hat{\vx}}_{t} &= h\!\left(\vF_t + \zeta_\mu\!\left(\hat{\vx}_t, \vu_t, \vy_t, \vF_t,  \zeta_E\!\left(\hat{\vy}_f\right)\right), \hat{\vx}_t\right).\label{hybridmodelwithencoder}
\end{align}
%Since we expect the DINOv2 features to include signals that are not relevant to the dynamics (lighting conditions, depth, etc.), we want to ensure that the mapping is distance-independent. 
During operation, the map will include features from various projection distances, requiring our network to effectively function within this distance-varying feature space. 
In order to train our network for these conditions, we collect features from $7$ different distance buckets, each spaced out by 10~m and at a maximum of 40~m in front of the vehicle. 
The exact projection distance is nominally in the range of $\pm5$~m but can have a large precentage of outliers due to the nature of the processing pipeline and environment.

The DINOv2 features change significantly based on distance.  
We have previously discussed the many issues that can cause the features to change, but all of these issues remain at runtime so we briefly discuss the impact they have on the features themselves. 
We ignore invalid values when computing the mean errors and see substantial $l_2$ differences in feature vectors as a function of projection distance. 
Even when looking at buckets that are next to each other in distance, we still see features varying by $30\%$ of the total range of values across the dataset for that specific feature.
Furthermore, occlusion causes substantial issues as a function of distance; we see the following occlusion percentages in distance buckets: $[1 , 0.87 , 0.92 , 0.92 , 0.84 , 0.65 , 0.42 , 0.21]$ for distances hindsight, -20m, -10m, ..., 40m. 
A negative distance means we have driven over the location and had the chance to collect data using the back LiDAR.
The decrease in validity in the closer distance buckets comes from changing ground plane conditions after the vehicle has compressed the traversed ground with a slower frame rate on the back camera.

We propose randomizing the projection distance by drawing upon features from any of the $7$ distance buckets or hindsight.
The randomized distance is maintained through the entire trajectory since varying it at each time step resulted in poor performance. 
We expect that the temporal dependence in invalid values is important for stable results when constrained to a single bucket of distance.
We augment the input features to the compression network with a flag $\{-1, 1\}$ to indicate if they are missing and replace the missing features with the mean of the training dataset.
%When a single input is nan the entire set of features are replaced with the mean features of the training dataset.

\section{Results}\label{results}

\begin{figure*}[t!]
\centering
\begin{subfloat}[Compression Type \label{fig:compression_type}]{
\centering
\includegraphics[page=5, trim={0cm, 0cm, 0cm, 0cm}, clip, width=0.31\textwidth]{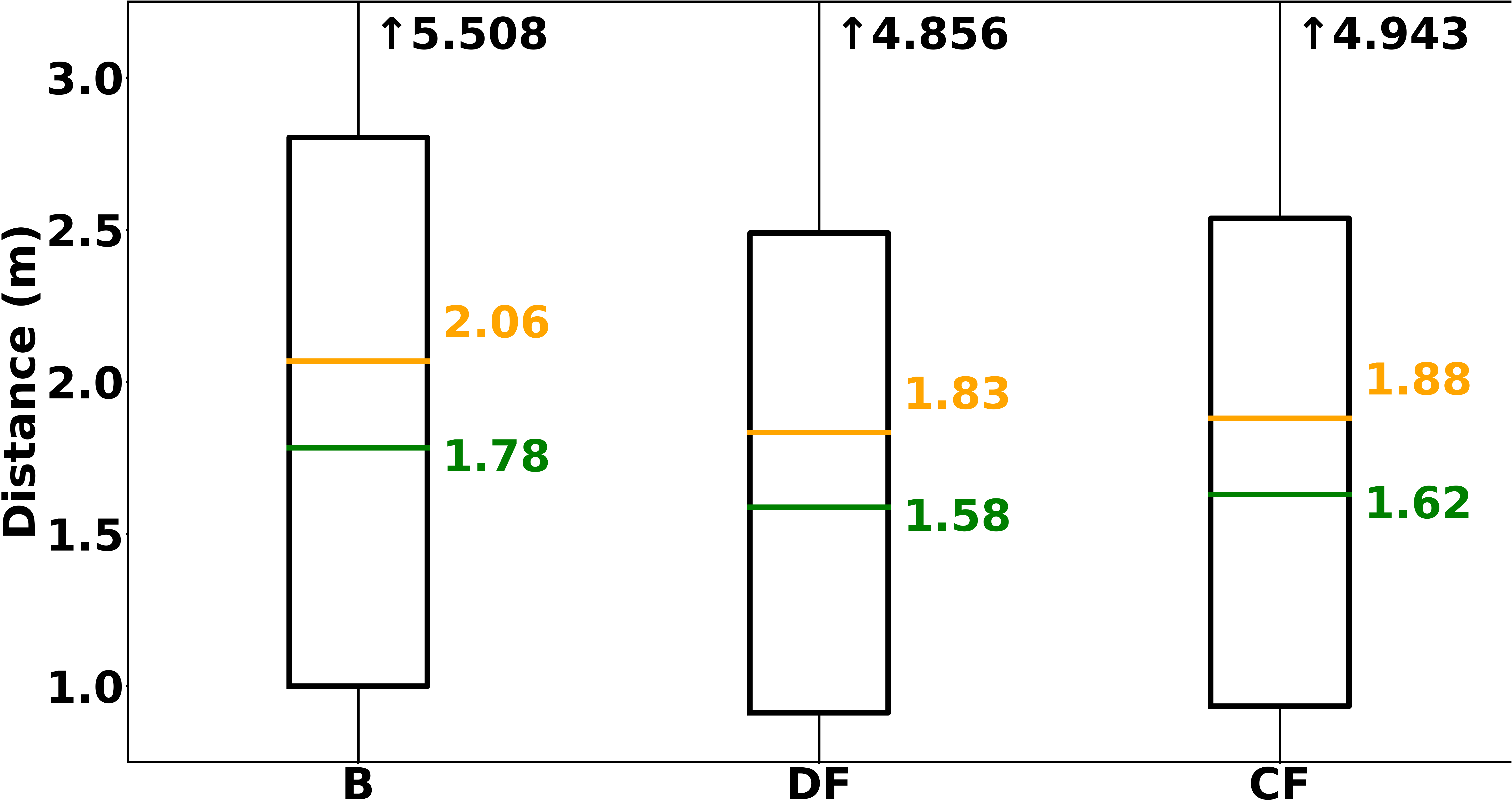}   }
\end{subfloat}
\begin{subfloat}[Compression Dimension \label{fig:compression_size}]{
\centering
\includegraphics[page=5, trim={0cm, 0cm, 0cm, 0cm}, clip, width=0.31\textwidth]{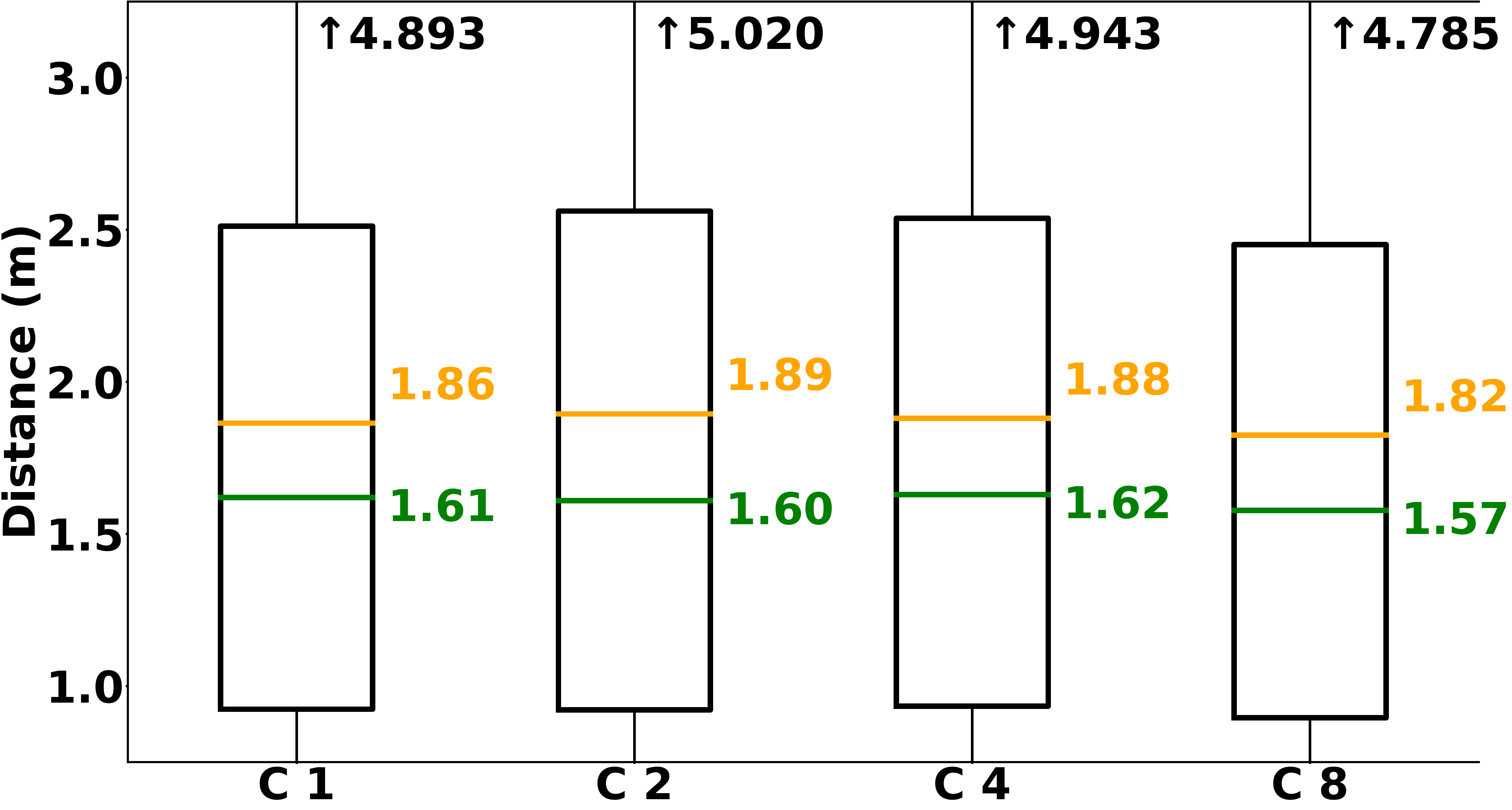}   }
\end{subfloat}
\begin{subfloat}[Number Of PCA Features \label{fig:num_pca_features}]{
\centering
\includegraphics[page=5, trim={0cm, 0cm, 0cm, 0cm}, clip, width=0.31\textwidth]{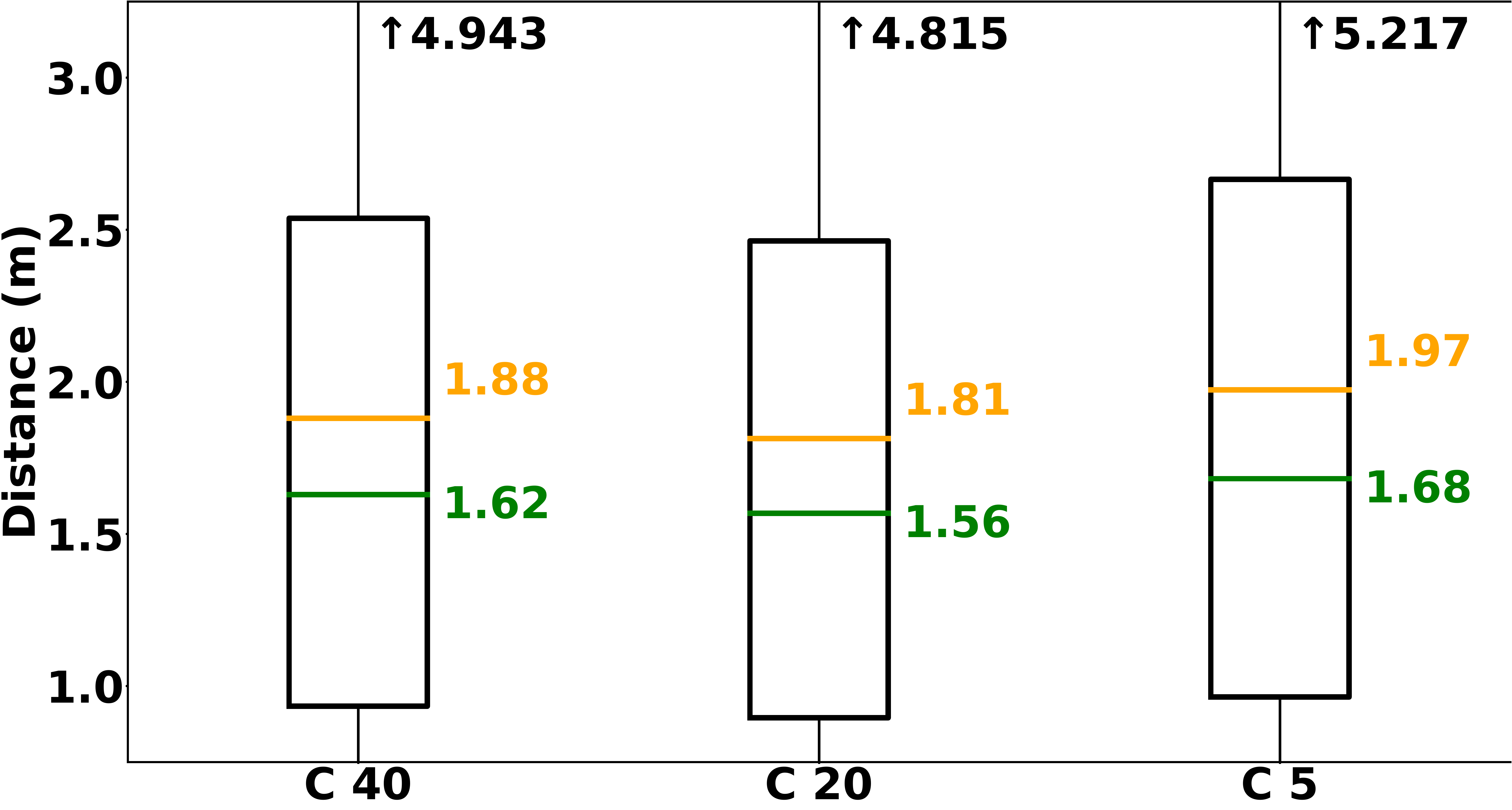}   }
\end{subfloat}
\caption{Distance error of models at $5s$ using best features in hindsight, B is a no feature baseline, DF is directly inputting features into the network, and C is compressing features. 
The model \textit{CF} in \ref{fig:compression_type}, \textit{C 4} in \ref{fig:compression_size}, \textit{C 40} in \ref{fig:num_pca_features} are all the same and axis are kept consistent between graphs.
Whiskers are defined as $\pm 1.5 IQR$ and given by the values with arrows, the green line defines the median and the orange the mean.
\ref{fig:compression_type} shows the different ways of inputting the features compared to no features.
\ref{fig:compression_size} shows the effect of changing the final compression size and the method is robust to this variable.
\ref{fig:num_pca_features} shows the impact of using a variety of PCA features.
}
\label{fig:hindsight_results}
\end{figure*}

Our results section is divided into two parts. 
First, we demonstrate the improvements to the dynamics modeling that terrain-based features can provide when the features are purely derived from hindsight.
Second, we show that our approach is able to handle features from farther distances when naive methods fail.

% TODO add in labels for images to refer back to
All networks are trained for 30 epochs on a dataset of $\approx 2M$ $5$-second trajectories with a test set of size $\approx 250k$ trajectories.
All summary statistics are computed at the end of the prediction horizon.
Initialization LSTMs have $20$ hidden layers, and the predictor LSTM uses a hidden size of $4$ with an additional output network that transforms the output dimension with a $20$ neuron hidden layer.
The training and test sets are derived from the same logs but have no overlap.
To model delays (brake, engine, steering), data is drawn from autonomous driving logs collected over a year from four distinct environments in three locations.
Most data (64\%) comes from the Mojave Desert near Helendale, CA, featuring loose sand and compressible creosote bushes (see Fig. \ref{varied_terrains} top middle \& right, bottom right).
The second environment (28\%) is from Halter Ranch near Paso Robles, CA (Fig. \ref{varied_terrains} bottom left) and includes dry grasses, oak woodland, and steep slopes up to $40^\circ$.
The final two environments are coastal dunes (\ref{varied_terrains} top left) (4\%) and coastal sage scrub (i.e., thick vegetation \ref{varied_terrains} bottom middle) (4\%) collected near Oceanside, CA.
From anecdotal experience, all environments bring about unique vehicle dynamics.

A major challenge across environments is ground plane estimation and occlusion at different distances.
Helendale's low vegetation density allows for easy ground plane estimation, while Halter Ranch's 1-2 ft tall grass makes makes that estimation difficult, often leading to invalid DINOv2 features.
Furthermore, the trees in Halter Ranch and dense vegetation in Oceanside prevent us from observing the occluded terrain before traversing past it.
These issues, while not present when using DINOv2 for short-horizon work (such as in \cite{magic_vfm}), create room for improvement, but we highlight how our method can handle features collected at various distances.

\subsection{Post-Processed Visual Features}
\label{sec:results_post_processed}

On the post-processed visual features, we see that networks with DINOv2 information as inputs outperform the model without any visual information as shown in Fig.~\ref{fig:compression_type}.
All models perform well since the median distance traveled by a trajectory in the dataset is $\approx 24 \text{m}$ with a median speed of $\approx 4.7 \text{m/s}$.
The compression scheme (CF) performs slightly worse than directly inputting the features into the network (DF), likely because of the time-dependence of visual features.
Note that naively inputting the (raw) features is not feasible in real time because of the mapping pipeline sensitivity to additional features.
Still there is a $\sim\!10\%$ decrease in mean summed loss with the compressed version compared as compared to the absence of DINOv2 features.
Most of the error reduction comes from $\sim\!8\%$ reduction in the mean error for $v_x$ at the end of the time horizon $5s$.
This directly impacts the accuracy for the yaw since the parametric yaw rate is heavily dependent on speed.

We vary the output dimension $n_\text{encoder}$ in Fig.~\ref{fig:compression_size} and vary the number of PCA features in Fig.~\ref{fig:num_pca_features} to check sensitivity to these parameters.
Smaller compression sizes $n_\text{encoder}$ significantly reduce the memory and computation requirements of the mapping pipeline, so the smallest vector necessary should be used.
We see similar performance as we vary these parameters, indicating that compression does not lose useful dynamical information.
The PCA feature sensitivity is irrelevant at runtime but can speed up post-processing.
Overall, our method is robust to these variations.

\subsection{Varying Distance Features}

\begin{figure}[t!]
\centering
\begin{subfloat}[Naive Evaluation On Larger Distances \label{fig:naive_distance_results}]{
\centering
\includegraphics[page=5, trim={0cm, 0cm, 0cm, 0cm}, clip, width=\columnwidth]{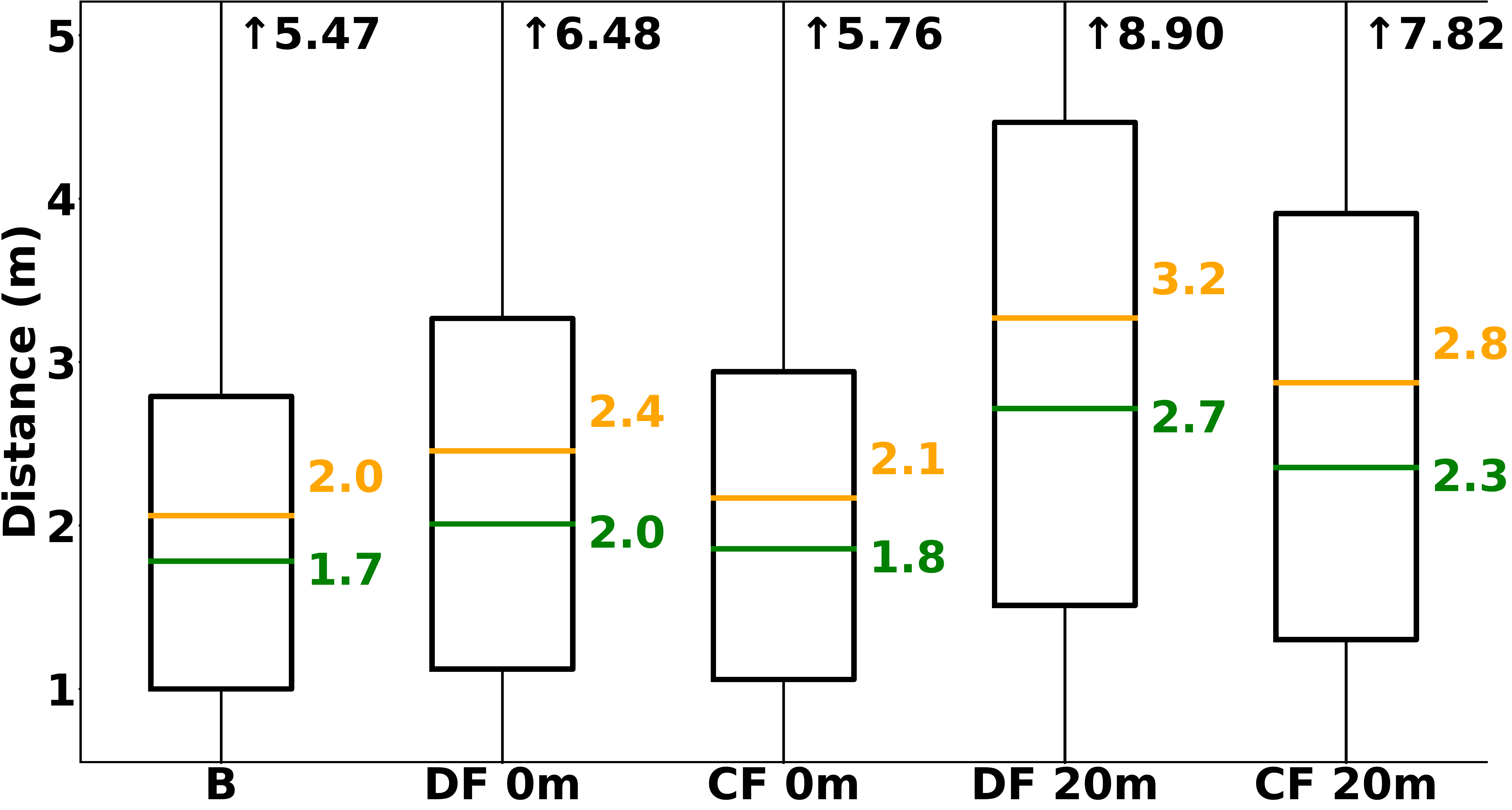}   }
\end{subfloat}
\begin{subfloat}[Proposed Approach (DC) On Larger Distances \label{fig:improved_dist_results}]{
\centering
\includegraphics[page=5, trim={0cm, 0cm, 0cm, 0cm}, clip, width=\columnwidth]{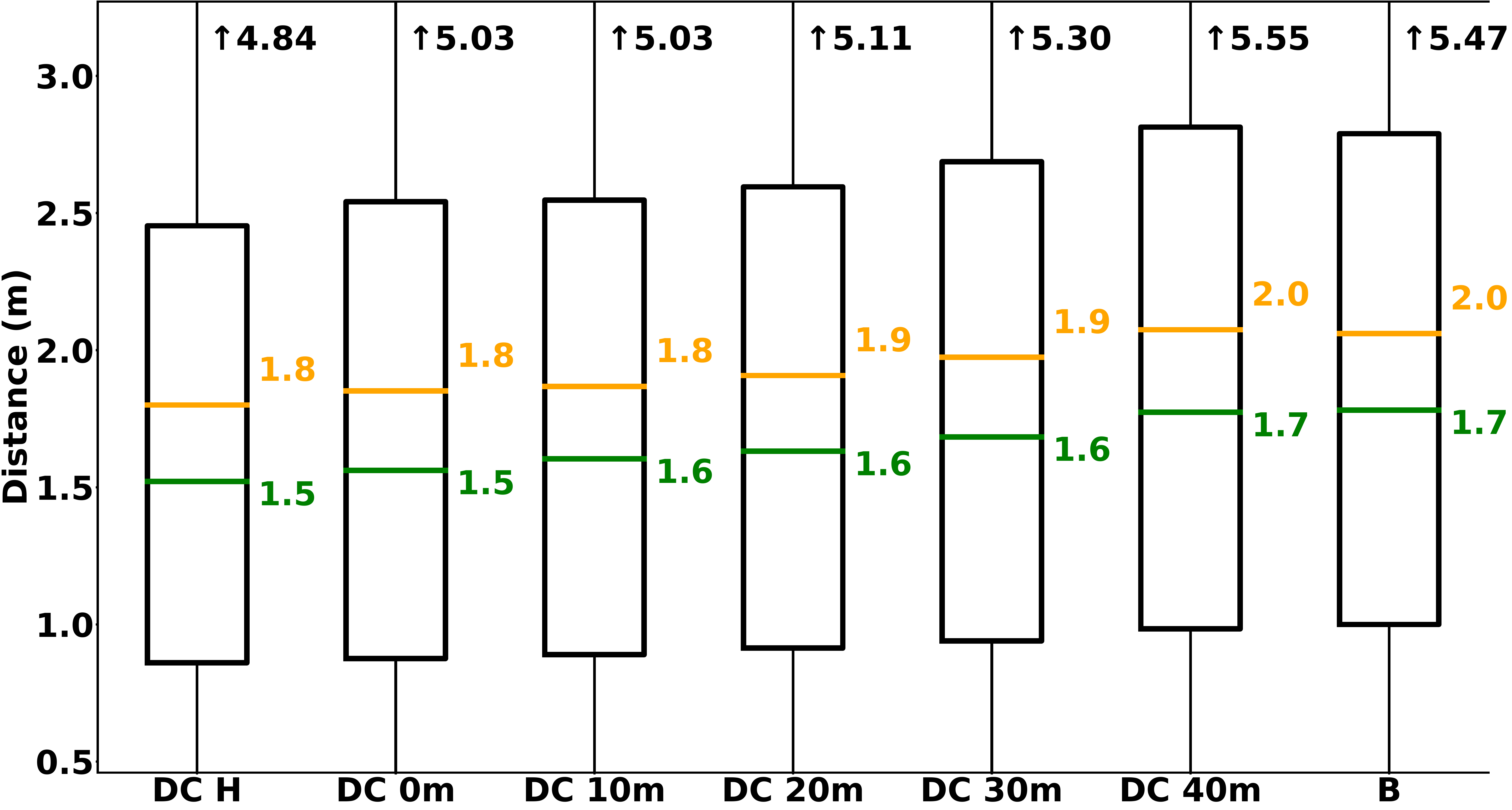}   }
\end{subfloat}
\caption{Distance error of at $5s$ models on features at varying projection distances. 
The model \textit{DF} and \textit{CF} are kept consistent from Fig.~\ref{fig:compression_type}.
\textit{DC} is our proposed distance independent approach.
Whiskers are defined as $\pm 1.5 IQR$ and given by the values with arrows, the green line defines the median and the orange the mean.
Fig.~\ref{fig:naive_distance_results} shows that using larger projection distances performs worse than a no-feature baseline \textit{B} with naive training methods.
Fig.~\ref{fig:improved_dist_results} shows that our approach can give improved results using visual features at realistic distances.
}
\end{figure}
In the previous section, we showed that hindsight features are able to improve the prediction accuracy, but these methods rely on hindsight feature information that does not exist at runtime. We next evaluate our models on features collected at distances more representative of those during vehicle operation, and we see that the networks trained on hindsight features fail in Fig.~\ref{fig:naive_distance_results}.
Plotting trajectories using the farther distances shows that the network is unable to adapt to the different distribution of features and has a tendency to generate nonsensical trajectories when tested on them.
Note that the compressed version (CF) is less sensitive to directly inputting them (DF).
We expect this is due to the compression learning a basis that is less sensitive to distance-dependent artifacts (even without any explicit penalty in training) and further motivates our use of this architecture.

Following the approach outlined in Section~\ref{sec:learning_dist_ind_features}, our approach (DC) is able to handle a variety of distances with lower error than the featureless baseline, as shown in Fig.~\ref{fig:improved_dist_results}.
When evaluated on hindsight data, training with naively-inputted features and distance-independent training perform similarly, meaning the distance-independent training is learning similar information just on a distance-independent basis.
At greater distances, there begins to be significant issues with LiDAR sparsity and occlusions that make the features unreliable, as seen in the high percentage of missing data (see Section~\ref{sec:learning_dist_ind_features}), which explains the lack of improvement using visual features beyond $30$m.
We focus our attention on the features in the range from $0 \text{m} \rightarrow 30 \text{m}$ since this is primarily where the dynamics predictions occur while driving the vehicle. Our method outperforms the featureless baseline at those distances.
%it's a tiny bit over 6 pages now
% we can fix it. looking twhere to cut same
% that did it

\section{Conclusion}

We present a hybrid model for vehicle dynamics that incorporates visual features of the terrain to anticipate changes in terradynamics. Our method improves upon the baseline model (dynamics learning without vision) by approximately 10\% without significant computational burden. We also provide analyses and ablations of the hyperparameters used in the feature compression network and into the dependence on the distance at which the visual feature is collected to mimic realistic driving conditions.% where the hindsight features are not available. % can we delete "where the hindsight features are not available"?
We test our method on an extensive dataset of driving data across various terrains.
Future work will explore further aspects such as the distance dependence of the visual features, occlusion handling, and ground plane estimation in vegetative environments.

\bibliographystyle{plain}
\bibliography{references}

\end{document}